\documentclass[letterpaper, 10 pt, conference]{ieeeconf}  
\usepackage{graphicx}
\usepackage{leftidx} 
\usepackage{amsmath}
\usepackage{xcolor}
\usepackage[linesnumbered,ruled]{algorithm2e}

\SetCommentSty{mycommfont}

\makeatletter

\IEEEoverridecommandlockouts                              
\title{\LARGE \bf
Neural Network-Based Collaborative Filtering for Question Sequencing
}

\author{Lior Sidi and Hadar Klein
\thanks{}
}

\begin{document}

\maketitle
\pagestyle{plain}

\begin{abstract}
E-Learning systems (ELS) and Intelligent Tutoring Systems (ITS) play a significant part in today's education programs. Sequencing questions is the art of generating a personalized quiz for a target learner. A personalized test will enrich the learner's experience and will contribute to a more effective and efficient learning process. In this paper, we used the Neural Collaborative Filtering (NCF) model to generate question sequencing and compare it to a pair-wise memory-based question sequencing algorithm - EduRank. The NCF model showed significantly better ranking results than the EduRank model with an Average precision correlation score of 0.85 compared to 0.8.
\end{abstract}

\section{INTRODUCTION}

E-Learning systems (ELS) and Intelligent Tutoring Systems (ITS) play a significant part in today's education programs. Learners can study from their laptop and gain education from top lectures with Massive Open Online Courses (MOOCs). Learners can access many different educational resources such as lectures, summaries, exercises, and exams. Recommender systems (RS) personalize ELS to learners in order to enrich their experience so they could learn more effectively and efficiently while keeping them motivated \cite{Klasnja-Milicevic2015RecommenderExtensions}. 

One of the many challenges that RS and ELS deal with is generating a personalized test for a target learner. The primary motivation behind personalized tests is to avoid the frustration of the learner from too easy or difficult questions under a certain context. In order to create a personalized test, the RS system should consider the learner's personalized difficulty, capabilities, context, learning styles, and habits.

In order to generate a personalized test, one must first assess the difficulty level of all the questions. The main article that this paper is based on addresses this issue with personalized sequencing of questions \cite{Segal2014}. The next stage of the personalized test is to decide the order of questions for the target learner.

Personalized sequencing in ELS is the learner's path through a collection of learning objects. Sequencing is an essential part of the Sharable Content Object Reference Model (SCORM), an e-learning software product standard, and is applied by different mechanisms such as schedule-based sequencing, artificial intelligence-based sequencing, collaborative learning or customized learning \cite{Klasnja-Milicevic2017E-LearningSystems}.

In the first section, "Related work" we overview different aspects of recommendation systems in E-learning and explain in detail the question sequencing task. We review the EduRank algorithm, a memory-based algorithm for question sequencing. In the "Method" section, we present a novel approach for sequence questions by using Neural Collaborative Filtering (NCF) model. The NCF has two main advantages over the EduRank memory-based algorithm. First, the NFC model can represent complicated connections between users and items. Second, It can be adaptively retrained by the user's feedback to tune the question sequencing. In the "Evaluation" section, we examine the different parameters in the NCF network and finally compare the optimized model with the traditional EduRank model on the Algebra dataset \cite{Koedinger2010}. In the "Results" section, we present the evaluation results. The NCF got significantly better results than the EduRank with an AP score of 0.85 compare to 0.8 and with a positive spearman rho ranking of 0.26. In the "Discussion" section, we interpret the results, and in the "Conclusions" section that we point out major conclusions and propose future work.

\section{Related Work}
\subsection{Recommendation Systems in E-learning}
A common approach to generate item rankings is to use Collaborative Filtering (CF) methods. Of these, most methods order items for target users according to their predicted ratings. Thai-Nghe et al, proposed to predict student performance using matrix factorization \cite{Thai-Nghe2011MatrixPerformance.}. Their methods address specific latent vectors that explain when a learner is guessing or when he made a true mistake. Furthermore, they used the tensor factorization algorithm to combine the learner's knowledge improvement over certain time-context. They applied their algorithms on the KDD 10 dataset \cite{Koedinger2010ADataShop}, and the results seemed promising compared to other classification methods.

Another CF approach relies on the similarity between item ratings of different users to directly compute the recommended ranking over items. Segal, Ktzir and Gal applied this approach and suggested the personalized pre-ordering of questions by difficulty using CF and social choice methods \cite{Segal2014EduRankE-learning}. A more detailed explanation of their study is presented in the next sub-section.
Rômulo, Direne and Marczal suggested an Adaptive Sequencing Method (ASM) algorithm for sequencing exercises for learners by difficulty \cite{Silva2015UsingPersonalization.}. Their strategy for sequencing the exercise starts with the easiest and finishes with the hardest exercise where they build a sequence of exercises as follows: if the student is correct, the next exercise can skip a certain step-size of questions ordered by difficulty. If the number of attempts in an exercise exceeds the average number of attempts, the next recommended exercise will be a mid-range difficulty. They tested their algorithm on 149 students, and the results showed that ASM increased the student score.

Tarus, Niu and Yousif suggested a generic environment for recommending different types of learning resources such as lectures, assignments, and exams \cite{Tarus2017AMining}. Their environment first creates an ontology on the domain and the learner's learning style and knowledge level. The system uses the learner's weblogs (Ip, time, HTTP page request) to generate the top-N resources using the Generalized Sequential Pattern (GSP) algorithm. The GSP finds frequent learning objects. Finally, the top-N items are ordered per user using Sequential Pattern Mining (SPM) that give importance weights for each learning item. Their hybrid approach showed significantly better performance and satisfaction for the learners. 

Wang, Wang and Yeung proposed using collaborative filtering with a deep neural network architecture for recommender systems \cite{HaoWangNaiyanWang2014CollaborativeSystems}. 
They combined ratings (sparse data) and auxiliary information such as item content information (can also be sparse) to generate new user-item ratings. Their collaborative deep learning model (CDL), preforms deep representation learning for the content information and collaborative filtering for the ratings (feedback) matrix. The CDL model's recall measure out-performed other matrix factorization and SVD like methods.

\begin{figure}[h]
\centering\includegraphics[width=1\linewidth]{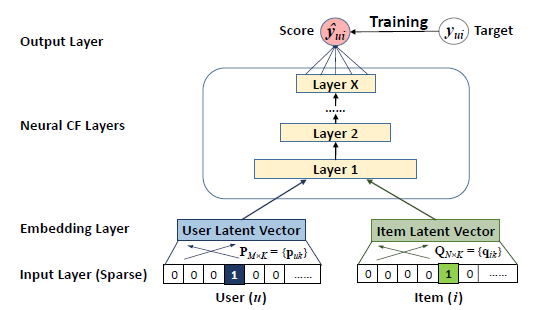}
\caption{Neural Network-based Collaborative Filtering Architecture}
\label{fig:NCF}
\end{figure}

He et al, presented the Neural network-based Collaborative Filtering framework (NCF) \cite{He2017NeuralFiltering}. As opposed to the previous work, they used the rating information without the content data. They proposed a model-based algorithm using two embedding layers of user latent factors and item latent factors. The concatenated product of these two embedding layers is then used as the first layer in a deep multi-layer neural architecture ("the neural collaborative filtering layers"), to map the latent vectors to prediction scores. The NCF model needs many training data to be accurate, but it can reveal complex latent features (more than the simple SVD model). Figure ~\ref{fig:NCF} outline the NCF architecture by \cite{He2017NeuralFiltering}.

\begin{figure}[h]
\centering\includegraphics[width=1\linewidth]{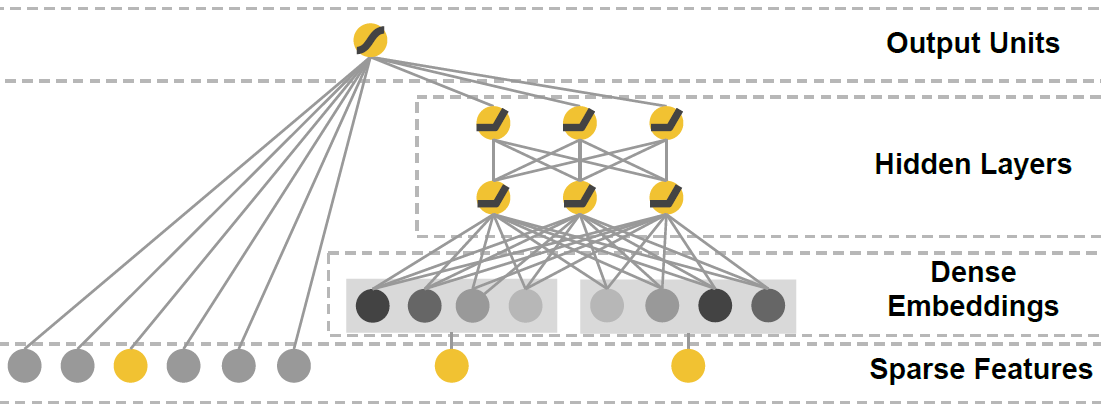}
\caption{The Wide (left side) and Deep (right side) model}
\label{fig:WDL}
\end{figure}

Cheng et al, presented the Wide and Deep Learning model for Recommender Systems (WDL) \cite{Cheng2016WideSystems}. The WDL model combines the generalization strength of deep neural networks with the memorization of feature interactions (through a wide set of cross-product feature transformations), which are effective and interpretable. The wide component is a generalized linear model, while the deep component is a feed-forward neural network. The WDL architecture can be seen in figure ~\ref{fig:WDL}.

\subsection{Question Sequencing - The EduRank study}
The EduRank study introduces an algorithm that uses collaborative filtering and social choice in order to produce a personalized ranking of new questions sequenced by their difficulty. Therefore, this study addresses the first step in generating a personalized test. The Edurank study presents some key concepts: 
\begin{itemize}
\item Partial order - Let $\succ_{j}$ be the partial order of a set of questions for student j. If $q_n \succ_{j} q_m$ then $q_n$ is more difficult than $q_n$ for student j.
\item NDPM - The Normalized Distance-based Performance Measure (NDPM) is used for evaluating a proposed system ranking to a reference ranking. It differentiates between correct orders of pairs, incorrect orders, and ties. NDPM is used once in the evaluation of the algorithm \cite{Segal2014EduRankE-learning}.
\item AP Rank Correlation - the Average Precision correlation metric (AP or SAP) is also used for evaluating a proposed system ranking to a reference ranking, but it gives more weight to errors over items that appear at higher positions in the reference ranking. AP is used in the EduRank algorithm and also during the evaluation of the algorithm  \cite{Segal2014EduRankE-learning}. 
\end{itemize}

\noindent The Edurank study used two real world educational datasets: 
\begin{enumerate}
\item The Algebra 1 dataset that was published in the KDD cup 2010 by the Pittsburgh Science of Learning Center (PSLC) \cite{Koedinger2010}. This dataset contains about 800,000 answering attempts by 575 students, collected during 2005-2006. The features extracted for each question were: question ID, the number of retries needed to solve the problem by the student, and the duration of time required by the student to submit the answer.
\item The K12 unpublished dataset obtained from an e-learning system installed in 120 schools and used by more than 10,000 students. This dataset contains about 900,000 answering attempts in various topics, including mathematics, English as a second language, and social studies. A domain expert ranked the questions with a difficulty score (CER) between one to five.  
The features extracted for each question were: question ID, the answer provided by the student, the associated grade for each attempt to solve the question, CER score, mastery level of the student on the question's topic (TBR).
\end{enumerate}

In the EduRank Algorithm, every question has a difficulty degree comprising of the first attempt grade and the number of retries. For the PLSC dataset, the elapsed time solving the question was also considered. The Edurank model is a memory-based algorithm because it needs to save all the user's similarities in order to generate a rating, the difficulty score. In algorithm 1 we present EduRank algorithm. 

\begin{algorithm} [h]
	\caption {The original EduRank Algorithm}
    \SetKwInOut{Input}{Input}
    \SetKwInOut{Output}{Output}
	
    \Input{
    {Set of students S.}
    \newline {Set of questions Q.}
    \newline {For each student $s_j \in S$, a partial ranking $\succ_j$ over $T_j \subseteq Q$.}
    \newline {Target student $s_i \in S$.}
	\newline {Set of questions $L_i$ to rank for $s_i$.}
    }

    \Output{a partial order $\hat{\succ_i}$ over $L_i$.}

	\ForEach{$q\in L_i$}{
    $c(q)={\sum\limits_{q_l\in{L\setminus{q}}}}{rv(q,q_l,S)}$
    
    }
	$\hat{\succ_i} \gets {\forall(q_k,q_l) \in \binom{L_i}{2}, q_k \hat{\succ_i} q_l \text{ iff } c(q_k)> c(q_l) }$
	
    \Return {$\hat{\succ_i}$}
\label{alg:algorithm1}
\end{algorithm}

The EduRank algorithm uses the known partial ranking $\succ_{j}$ for each student $s_{j}$ over the group of questions which that student has answered ($T_{j}$). The output of the algorithm is the partial order over the group of questions that a target student ($s_{i}$) has not yet answered ($L_{i}$). For each question $q$ in $L_{i}$ a Copland score is calculated ($c(q)$). the Copland score is a representative of the difficulty rank of $q$ in $L_{i}$, and it is described in equation~\ref{eq:copland}.

\begin{equation} 
c(q)={\sum\limits_{q_l\in{L\setminus{q}}}}{rv(q,q_l,S)}
 \label{eq:copland}
\end{equation}

\noindent $rv(q,q_l,S)$ is the aggregated voting of the order of $(q,q_l)$ amongst all of $s_i$'s neighbors. $rv(q,q_l,S)$ is described in equation~\ref{eq:rv}. The aggregated voting can be perceived as a competition between $q$ and $q_l$.

\begin{itemize}
\item $q$ beats $q_l$ if the number of wins of $q$ over $q_l$  computed over all of $s_i$'s neighbors is higher than the  number of loses. In this case $rv(q,q_l,S)=1$. 
\item If the opposite occurs then $rv(q,q_l,S)=-1$.
\item and if the number of wins equals the number of losses then $rv(q,q_l,S)=0$.
\end{itemize}

\begin{equation} 
rv(q,q_l,S)=sign({\sum\limits_{j=S\setminus{i}}}{s_{AP}	(T_{i},\succ_{i},\succ_{j})}\cdot \gamma (q,q_l,\succ_{j})
 \label{eq:rv}
\end{equation}

\noindent The  neighbor $j$'s win or loss of $q$ over $q_l$ is expressed by equation~\ref{eq:wins}

\begin{equation} 
\gamma (q,q_l,\succ_{j})= 
						\begin{cases}
                           1  &  \text{if  } q \succ_{j} q_l \\
						   -1  &  \text{if  } q_l \succ_{j} q \\
                           0 &  \text{otherwise} 
                        \end{cases}
 \label{eq:wins}
\end{equation}

\noindent Every neighbor's win or loss of $q$ over $q_l$ is normalized by a similarity measure between that neighbor and $s_i$. The similarity measure is notated as $s_{AP}	(T_{i},\succ_{i},\succ_{j})$. It is based on the similarity between $s_i$ and $s_j$ regarding $s_i$'s known partial ranking $\succ_{i}$ over the group of question which have been already answered ($T_{i}$). The similarity score is based on an AP Rank Correlation metric where disagreements over questions that are perceived more difficult are more heavily penalized.

The EduRank algorithm's performance was compared to other ranking algorithms using the NDPM and AP scores. The other algorithms were CER, TBR, a KNN method using the Pearson correlation (denoted UBCF), a matrix factorization method using SVD (denoted SVD), and EigenRank, which are all explained in \cite{Segal2014}. EduRank outperformed all the other algorithms. The EduRank algorithm is simple, and due to its' collaborative filtering feature, the execution time is near the UBCF and better than SVD and EigenRank. 
The EduRank Algorithm does not acquire any user intervention in order to inquire the finalized ranking for the unseen question and keeps that list of unseen questions for future use. 
The EduRank algorithm creates a static ranking for a target student with given known answers. In order to calculate new difficulty scores based on the student's new answers, the algorithm is recomputing all similarity measures and partial rankings on the entire dataset. Edurank is a memory-based algorithm which makes it hard to use on real-world high scaled datasets without compromising run time performance. We propose using deep collaborative methods that are model-based to speed up the prediction of new difficulty scores based on user interaction.

\section{Method - Neural Collaborative Filtering}
Neural network-based Collaborative Filtering (NCF) captures the user to item interaction with a deep learning model by replacing the inner product of the matrix factorization with a neural network architecture \cite{He2017NeuralFiltering}.

As presented in Figure~\ref{fig:NCF}, the NCF is a multilayer network:
\begin{enumerate}
\item The inputs are the users and the items as two separate one-hot vectors.
\item Each input vector is connected to an embedding layer that functions as a latent factor vector. 
\item The embedding layers are then joined together to by concatenation to form the first layer in the neural collaborative filtering layers (stacked layers of weights and neurons).
\item the final layer is connected to one output neuron, which predicts the difficulty score.
\end{enumerate}

\begin{algorithm} [h]
	\caption {The NCF Algorithm}
    \SetKwInOut{Input}{Input}
    \SetKwInOut{Output}{Output}
	
    \Input{
    {Set of students S.}
    \newline {Latent factor size k}
    \newline {Number of hidden layers l (depth)}
    \newline {activation functions A}
    \newline {Set of questions Q.}
    \newline {Set of known difficulty scores D}
    \newline {Target student $s_i \in S$.}
	\newline {Set of questions $L_i$ to rank for $s_i$.}
    }

    \Output{a partial order $\hat{\succ_i}$ over $L_i$.}
	
    $newNCF \gets NCF(A,l,k) $  \tcp*{Create NCF model with l hidden layers and the wanted activation functions A.}    
	
    $newNCF.Fit(S,Q,D)$ \tcp*{Train NCF model on all known difficulty scores D.} 

	\ForEach{$q\in L_i$}{
    $d(i,q)= newNCF.predict(s_i,q)$  \tcp*{predict the difficulty score for student i and question q} 
    }
	
    $\hat{\succ_i} \gets {\forall(q_k,q_l) \in \binom{L_i}{2}, q_k \hat{\succ_i} q_l \text{ iff } d(i,q_k)> d(i,q_l) }$
	
    \Return {$\hat{\succ_i}$}
\label{alg:NCF}
\end{algorithm}

As presented in algorithm 2, the network trains on user-item previous rankings. Then new items for a target user can be ranked. The users are students, and the items are questions. The user-item ratings are the personalized questions difficulty scores. 

In order to use the NCF model for question sequencing, we train the network on the students' previous answers and their pre-calculated user difficulty score. For ranking new questions, we apply the model on the user and questions we wish to rank and get the predicted difficulty rank for each question. In the final step, questions are sorted by their difficulty and presented to the student.

One of the most critical aspects of the NCF model is the architecture configuration, such as the activation function, the latent factor size (width), the amount of the CF layer stacking (depth). In the evaluation section, we present the model parameters importance and finally compare the optimized model with the Edurank memory-based algorithm. 

\section{Evaluation}
We evaluated the models based on the same dataset used in the EduRank study, the Algebra 1 dataset. For the evaluation stage we filtered 250,000 question's answers and evaluate the question sequencing algorithms on 4 random questionnaires from 3 different users.

In this paper we conducted two types of experiments and evaluations:
\begin{itemize}
\item NCF parameter tuning:
\begin{enumerate}
\item Wide vs. Deep -  The wide architecture comprises of the concatenated embedding layers and an output neuron. The Deep architecture comprises of $l$ neural collaborative filtering layers in between the concatenated embedding layers and the output neuron. We tried $l = 1,2,4,8$.
\item Embedding size $k$ - We evaluated $k$ between 20 and 80.
\item Activation function $A$ - We evaluated $A = tanh, linear, relu$.
\end{enumerate}

\item Evaluation of the optimized NFC ranking vs. the EduRank ranking. We compared the EduRank algorithm with a memory size of the five most similar students to an optimized NCF algorithm. We used the true difficulty ranking as the referenced ranking and compared the EduRank and NCF to it. 
The evaluation metrics used were the average correlation score (SAP) and the Spearman's rho score (SR). The SAP is in the evaluation of the EduRank algorithm \cite{Segal2014EduRankE-learning}. The Spearman's rho score is a statistical correlation measure between two ranked variables \cite{Shani2011EvaluatingSystems}. Spearman's rho value is a continuous value between 1 and -1 where 1 is a perfect ranking correlation between the predicted ranking and the referenced ranking. The Spearman's rho calculation can be seen in equation~\ref{eq:sr}, where $d_i$ is the difference between the two ranks of each observation, an $n$ is the number of observations.

\begin{equation} 
SR={1-\frac{6\sum\limits_{i}{d_i^2}}{n(n^2-1)}}
\label{eq:sr}
\end{equation}

\end{itemize}

\section{Results}
\subsection{NCF tunning}
We started with evaluating the NCF architecture parameters. We set the architecture with fixed configurations of 1024 batch size and 0.25 for dropout rate. the training of the model is optimized with 'Adadelta' optimizer with a mean square error loss function.

\begin{figure}[h]
\centering\includegraphics[width=1\linewidth]{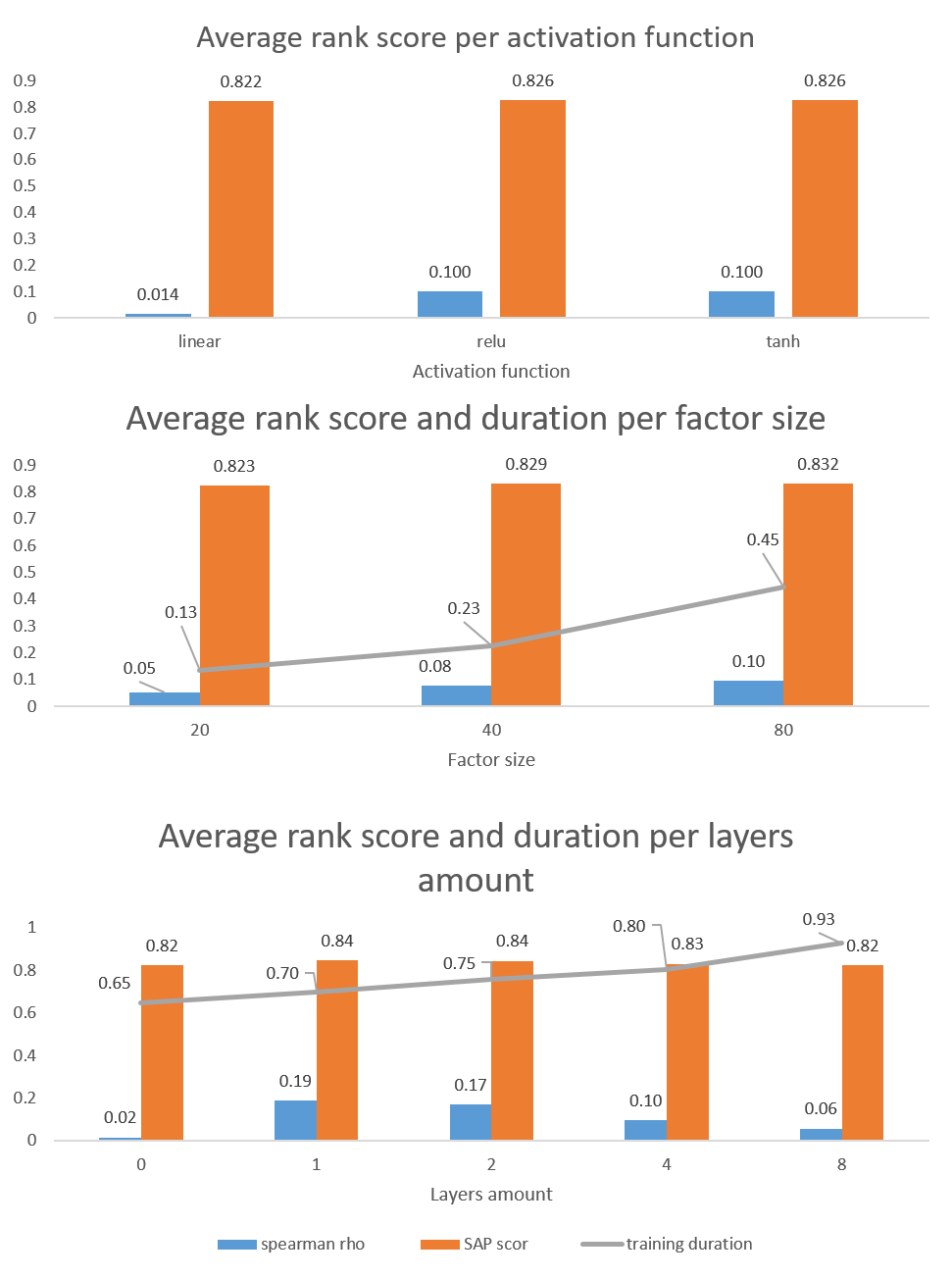}
\caption{NCF Parameters Tuning}
\label{fig:NCF_param}
\end{figure}

In Figure~\ref{fig:NCF_param} we present the parameters tuning results and evaluation. The X axis is the parameter value we evaluated and the Y axis is the ranking results (Spearman's rho and SAP), In the two bottom graphs there is also a gray line that represent the training duration per value. 

The top graph evaluates different activation functions with $l=4$ and $k=32$. As presented in the graph, the 'relu' and 'tanh' activation functions (SAP=0.825, SR=0.1) outperform the simple 'linear' function (SAP=0.822,SR=0.014). We choose the 'tanh' for the rest of the evaluation due to it's slightly higher AP rank and shorter fit and rate duration.

The middle graph evaluates $k$, the factoring size in the embedding layer (width sizes). As expected, higher $k$ values correlate with better ranking. The training duration also increases substantially when increasing the factor size of $k$. Therefore we choose a cost-effective size of $k=40$ that combines performance (low training duration) and high ranking results.

The bottom graph evaluates the network depth $l$, the number of stacked CF layers. Surprisingly deeper networks with more than one layer had lower Spearman's rho score and could not reach the 0.84 SAP score. The wide network with no CF layers (a simple factorized model) showed the worst results with only 0.02 Spearman's rho score. Therefore, the usage of at least one CF layer is necessary.

To conclude, the final optimized NCF network parameters are $k=40$, $l=1$, and 'tanh' as the activation function.

\subsection{NCF and EduRank comparison}
\begin{figure}[h]
\centering\includegraphics[width=1\linewidth]{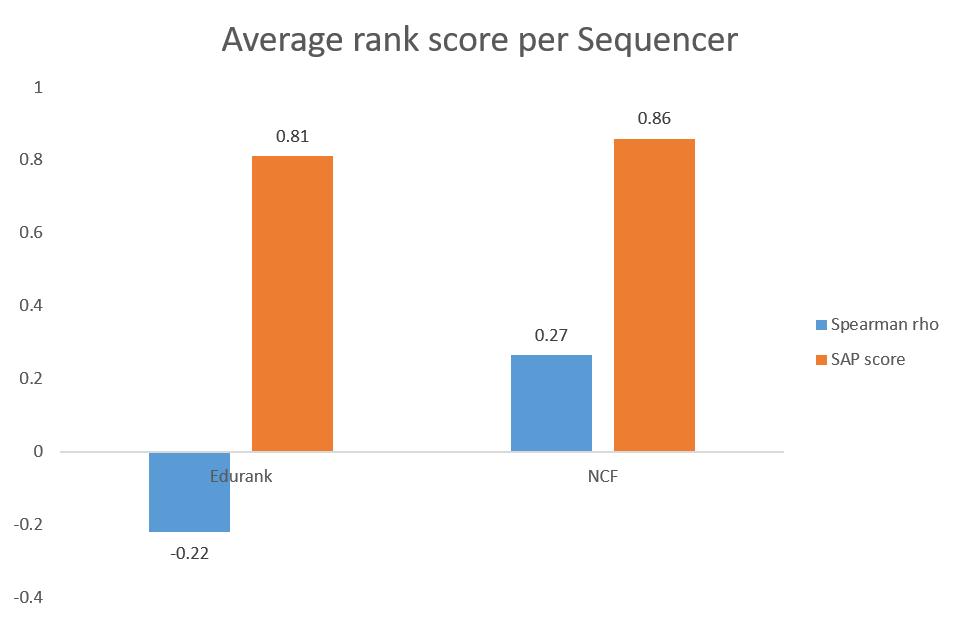}
\caption{rank results for EduRank and the CNF algorithms}
\label{fig:EduVsNCF}
\end{figure}

We compared the optimized CNF model with the known EduRank model. As demonstrated in Figure~\ref{fig:EduVsNCF}, the CNF model outperforms the EduRank model with a mean 0.86 SAP score and 0.27 Spearman rho compared to 0.81 and -0.22 in the EduRank algorithm. 

\begin{figure}[h]
\centering\includegraphics[width=1\linewidth]{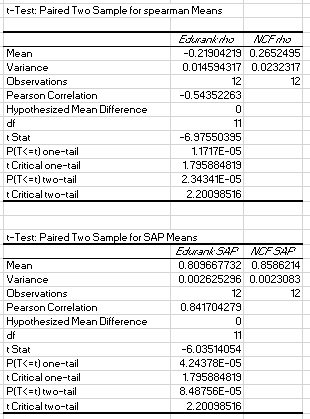}
\caption{Pairwise t-test results between EduRank and NCF ranking results}
\label{fig:ttest}
\end{figure}

To test the significance of the results, we used a pairwise t-test on the student-questionnaire results (total of 12 cases, three students per four questionnaires). In  Figure~\ref{fig:ttest}, we present the results of the test, in both tests for the Spearman and AP score the t-statistic is higher than the absolute value of the t-critical meaning that we reject the null hypothesis and the NCF rankings are significantly bigger then Edurank's.

\section{Discussion}

The optimized NCF model scored higher (SAP=0.86, SR=0.27) than the EduRank model (SAP=0.81, SR= -0.22). Deep learning showed good results compare to the memory-based algorithm under real-world circumstances where the memory size of the model is limited. Therefore we recommend the optimized NCF model to evaluate unseen personalized question difficulty scores.

Evaluating deep learning architectures is an endless game due to hyperparameter tuning.  In our evaluations, we addressed the most critical parameters that affect the NCF network: the embedding size ($k$, "width") and the number of hidden CF layers ($l$, "depth"). We encountered a tradeoff between increasing the width and gaining better ranking scores on the one hand and decreasing the performance (longer training time) on the other hand. We chose a cost-effective $k$ value that considers both performance and ranking scores. 
Regarding the depth size, we were surprised to see a decrease in the ranking scores as the number of hidden layers, $l$, grew. This result might have been a side effect of the vanishing gradient syndrome of deep neural networks. We propose more experiments with different activation functions (such as Relu) to test this theory.

The original Edurank algorithm is in Java Mahout (map-reduce), the implementation in this paper run locally with python and also demand many code optimization and parallelization to work in reasonable throughput.

One of the main advantages of the NCF on the memory-based algorithm is that due to its smaller size and efficient retraining, this advantage will allow deploying the NCF model in a real environment. 

\section{Conclusion}
In this paper, we used a neural collaborative filtering (NCF) model to sequence questions by difficulty and optimized it accordingly. We compared the NCF model with the EduRank algorithm with limited memory size, a memory-based algorithm for question sequencing. The NCF model showed significantly better results in ranking questions than the EduRank algorithm. 

The NCF has two additional advantages over the EduRank memory-based algorithm. First, the NCF model can represent complicated connections between users and items. Second, It can be adaptively retrained by the user's feedback to tune the question sequencing. The retraining of the NCF model can be done in batches and requires less computational complexity than the EduRank model.

We believe the next natural step should be to investigate the combination of a deep and wide architecture as proposed in \cite{Cheng2016WideSystems}. 
There is much more to investigate and research in this domain; it will be wise to use different types of architectures, such as recurrent neural networks, to recommend a future sequence of questions. It is also interesting to add content-based features to the entire network and checks any influence on the rating.

As mentioned, it is possible to use the NCF as a part of an online experiment where the model is retrained on the user's feedback.

\bibliographystyle{alpha}
\bibliography{main.bib}

\addtolength{\textheight}{-12cm}   


\end{document}